\documentclass[10pt,twocolumn,letterpaper]{article}

\usepackage[pagenumbers]{iccv}              % To produce the CAMERA-READY version
\usepackage{amsmath}
\usepackage{amssymb}
\usepackage{enumitem}
\usepackage{natbib}
\usepackage{graphicx}
\usepackage{float}
\usepackage{physics}
\usepackage{nth}
\usepackage{soul}
\usepackage{xcolor}

\usepackage[ruled,vlined]{algorithm2e}
\definecolor{iccvblue}{rgb}{0.21,0.49,0.74}
\usepackage[pagebackref,breaklinks,colorlinks,allcolors=iccvblue]{hyperref}

\def\paperID{13} % *** Enter the Paper ID here
\def\confName{ICCV}
\def\confYear{2025}

\title{Bayesian Multi-Scale Neural Network for Crowd Counting}

\author{Abhinav Sagar\\
University of Maryland, College Park, Maryland\\
College Park, Maryland\\
{\tt\small asagar@umd.edu}
}

\begin{document}

\maketitle

\begin{abstract}

Crowd counting is a challenging yet critical task in computer vision with applications ranging from public safety to urban planning. Recent advances using Convolutional Neural Networks (CNNs) that estimate density maps have shown significant success. However, accurately counting individuals in highly congested scenes remains an open problem due to severe occlusions, scale variations, and perspective distortions, where people appear at drastically different sizes across the image. In this work, we propose a novel deep learning architecture that effectively addresses these challenges. Our network integrates a ResNet-based feature extractor for capturing rich hierarchical representations, followed by a downsampling block employing dilated convolutions to preserve spatial resolution while expanding the receptive field. An upsampling block using transposed convolutions reconstructs the high-resolution density map. Central to our architecture is a novel Perspective-aware Aggregation Module (PAM) designed to enhance robustness to scale and perspective variations by adaptively aggregating multi-scale contextual information. We detail the training procedure, including the loss functions and optimization strategies used. Our method is evaluated on three widely used benchmark datasets—ShanghaiTech, UCF-CC-50, and UCF-QNRF—using Mean Absolute Error (MAE) and Mean Squared Error (MSE) as evaluation metrics. Experimental results demonstrate that our model achieves superior performance compared to existing state-of-the-art methods. Additionally, we incorporate principled Bayesian inference techniques to provide uncertainty estimates along with the crowd count predictions, offering a measure of confidence in the model's outputs.
 
\end{abstract}

\section{Introduction}

Crowd counting has garnered significant interest in the computer vision community in recent years due to its broad range of practical applications. These include estimating the number of people in political rallies, public demonstrations, concerts, religious gatherings, and sporting events. Moreover, the underlying methodologies can be adapted to related tasks such as counting cells in microscopic images, vehicles in aerial or satellite imagery, and animals in ecological monitoring. Despite its utility, crowd counting—particularly in highly congested scenes—remains a challenging problem. Two major factors contribute to this difficulty: (1) severe occlusions, clutter, and overlaps between individuals, and (2) large variations in scale and appearance due to perspective distortion, where individuals closer to the camera appear much larger than those farther away.

A wide range of algorithms has been proposed to address these challenges. The dominant paradigm in recent years involves the use of Convolutional Neural Networks (CNNs) combined with density map estimation. These methods predict a continuous density map over the input image, which, when integrated, yields the total object count. Training datasets typically provide only point annotations—often marking the head center of each individual—rather than full object labels or bounding boxes. This sparse annotation presents additional challenges for model learning. Earlier approaches to crowd counting relied on object detection or instance segmentation techniques to identify and count individuals. However, these methods proved inefficient and inaccurate, particularly in dense crowd scenarios, due to their high computational cost and poor performance under occlusion.

To address these limitations, regression-based methods were introduced. These approaches bypassed explicit object detection by learning a direct mapping from image features to a global count value. While this reduced the impact of occlusion and overlapping individuals, it failed to capture spatial information and struggled with varying object scales due to perspective effects. The most successful evolution in this field has been the use of density map estimation techniques. These models generate a density map that not only encodes the presence of individuals but also preserves spatial and scale information. By learning a mapping from the input image to a corresponding density distribution, these methods effectively handle both occlusion and scale variation. As a result, density map-based CNN approaches have become the state-of-the-art standard in modern crowd counting.

Bayesian techniques have emerged as powerful tools in deep learning-based crowd counting, offering not only accurate predictions but also principled uncertainty estimation. Unlike traditional deterministic models that provide point estimates, Bayesian approaches model the predictive distribution, enabling the quantification of both aleatoric uncertainty (inherent data noise) and epistemic uncertainty (model uncertainty due to limited data). This is particularly valuable in high-stakes applications such as surveillance and public safety, where understanding the confidence of a prediction is as important as the prediction itself. In the context of crowd counting, Bayesian neural networks can be implemented using methods such as Monte Carlo Dropout, variational inference, or deep ensembles to approximate posterior distributions. These techniques help mitigate overfitting, improve generalization, and offer uncertainty-aware predictions that can be leveraged for downstream tasks like active learning, anomaly detection, or dynamic resource allocation in real-time systems. Incorporating Bayesian inference thus adds a crucial layer of reliability and interpretability to modern crowd counting models.

\section{Related Work}

Several important contributions have advanced the field of crowd counting using deep learning techniques. One of the pioneering works in this domain is by \cite{zhang2015cross}, which introduced a cross-scene crowd counting approach using a switchable learning strategy that simultaneously optimized two objectives: crowd density estimation and overall count regression. Building on this idea, \cite{sam2017switching} proposed an end-to-end trainable switching CNN architecture that automatically selects the most suitable regressor for different crowd regions, improving robustness across varying densities. The concept of using multi-column architectures to capture features at different receptive fields was popularized by \cite{zhang2016single}, who replaced fully connected layers with $1 \times 1$ convolutional layers to reduce parameters while maintaining spatial resolution. Similarly, \cite{boominathan2016crowdnet} combined deep and shallow CNNs to capture both low-level and high-level features, enhancing performance in scenes with significant scale variation and occlusion.

\cite{ranjan2018iterative} introduced an iterative refinement approach where one CNN estimates a coarse density map, which is then progressively refined in a second stage. Meanwhile, \cite{sindagi2017cnn} proposed a multi-task cascaded CNN that jointly learns crowd count classification and density map regression, allowing shared feature learning and improved generalization. Further innovations include the multi-scale contextual encoding approach of \cite{liu2019context}, which explicitly models perspective distortion and demonstrates the benefit of multi-scale features in handling scale variation. \cite{zhang2018crowd} proposed a scale-adaptive fusion method that concatenates features extracted at different resolutions, while \cite{shang2016end} integrated local and global contextual information for predicting counts at multiple levels.

To incorporate top-down scene semantics, \cite{sam2018top} employed a feedback mechanism to refine predictions based on global scene context. \cite{shi2019revisiting} tackled the perspective challenge using perspective maps encoded as adaptive weighting layers to combine density predictions at multiple scales. \cite{jiang2019crowd} further explored scale adaptation with a multi-scale encoder and multi-path decoder framework for high-fidelity density map generation. Hybrid approaches have also emerged. \cite{liu2018decidenet} fused detection and regression using an attention mechanism to switch between the two paradigms based on crowd density. \cite{liu2019crowd} introduced a local pattern consistency loss, improving fine-grained density estimation through region-level correlation modeling. Attention mechanisms were also employed in \cite{hossain2019crowd} to enable both global and local scale selection via soft attention.

Semi-supervised and unsupervised learning approaches have shown promise as well. \cite{sam2019almost} utilized an autoencoder to extract transferable features from unlabeled data, while \cite{cheng2019learning} focused on identifying pixel-level subregions with high prediction errors to guide learning. \cite{sindagi2019ha} introduced a hierarchical attention framework, combining spatial and global attention modules across multiple scales to enhance focus on relevant crowd regions. Recent research has increasingly focused on integrating uncertainty modeling into crowd counting. \cite{oh2020crowd} and \cite{ma2019bayesian} independently proposed Bayesian formulations for crowd counting that yield both point estimates and uncertainty quantification. Their models are capable of estimating both epistemic uncertainty (related to model confidence) and aleatoric uncertainty (inherent data noise), improving reliability in ambiguous or high-density scenarios. \cite{idrees2018composition} presented a unified composition loss that jointly supervises count, density, and localization tasks, pushing the boundary of multi-task learning in dense scenes.

We summarize our main contributions as follows:

\begin{itemize}

\item We propose a novel deep neural network architecture for crowd counting, built upon a ResNet-based feature extractor. Our model incorporates a downsampling module using dilated convolutions to preserve spatial resolution, and an upsampling module using transposed convolutions to reconstruct high-quality density maps.

\item We introduce a novel Perspective-aware Aggregation Module (PAM) that improves robustness to scale and perspective distortions by adaptively fusing multi-resolution features across the network.

\item We provide comprehensive implementation details, including the network architecture, optimization strategy, loss functions, and the evaluation protocol. Our method is evaluated on three widely used benchmarks—ShanghaiTech, UCF-CC-50, and UCF-QNRF—using standard MAE and MSE metrics.

\item Our model achieves state-of-the-art accuracy while significantly reducing the number of parameters compared to existing methods. Furthermore, we incorporate principled Bayesian inference to estimate both epistemic and aleatoric uncertainties, making our system more interpretable and reliable for real-world deployment.

\end{itemize}

\section{Proposed Method}

\subsection{Dataset}

Experimental evaluations are conducted using three widely used crowd counting datasets: ShanghaiTech part A
and part B, UCF-CC 50, and UCF-QNRF. These datasets are described as follows:

\begin{itemize}

\item ShanghaiTech is made up of two datasets labeled as part A and part B. In Part A, there are 300 images for training and 182 images for testing, while Part B has 400 training images and 316 testing images. Most of the images are of very crowded scenes, such as rallies and large sporting events. Part A has a significantly higher density than Part B.

\item UCF-CC-50 contains 50 gray images with different resolutions. The average count for each image is 1,280, and the minimum and maximum counts are 94 and 4,532, respectively.

\item UCF-QNRF is the third dataset used in this work, which has 1535 images with 1.25 million point annotations. It is a challenging dataset because it has a wide range of counts, image resolutions, light conditions, and viewpoints. The training set has 1,201 images, and 334 images are used for testing.

\end{itemize}

\subsection{Network Architecture}

Our proposed network architecture consists of three primary modules: a feature extraction block, a reconstruction (upsampling) block, and a multi-head prediction module for density estimation and uncertainty quantification.

The feature extraction block is built upon a modified ResNet backbone enhanced with dilated (atrous) convolutions, which serve as the downsampling mechanism. Unlike traditional max-pooling or stride-based downsampling, dilated convolutions allow the receptive field to expand without sacrificing spatial resolution. This design is particularly effective in crowd-counting scenarios where objects (i.e., people) appear at varying scales due to perspective distortion. By capturing multi-scale contextual information, the dilated ResNet-based encoder mitigates issues related to severe occlusion and scale variation.

Following the encoder, the reconstruction or upsampling block utilizes transposed convolutional layers (also known as deconvolutions) to progressively restore the spatial resolution of the feature maps. To preserve fine-grained details lost during encoding, skip connections are introduced between corresponding encoder and decoder layers. These lateral connections form a U-Net-like structure, facilitating efficient gradient flow and enabling the network to fuse low-level and high-level information.

The final part of the architecture is the multi-head output module, which consists of three branches:

\begin{itemize}

\item The density map head, which produces a high-resolution density map. When integrated spatially, this map yields the total estimated count of people in the input image.

\item The epistemic uncertainty head, which estimates uncertainty arising from model limitations, is approximated via the Monte Carlo dropout technique.

\item The aleatoric uncertainty head, which models noise inherent in the input data, is particularly relevant in cluttered or poorly illuminated scenes.

\end{itemize}

\begin{figure}[htp]
\centering
\includegraphics[width=8cm]{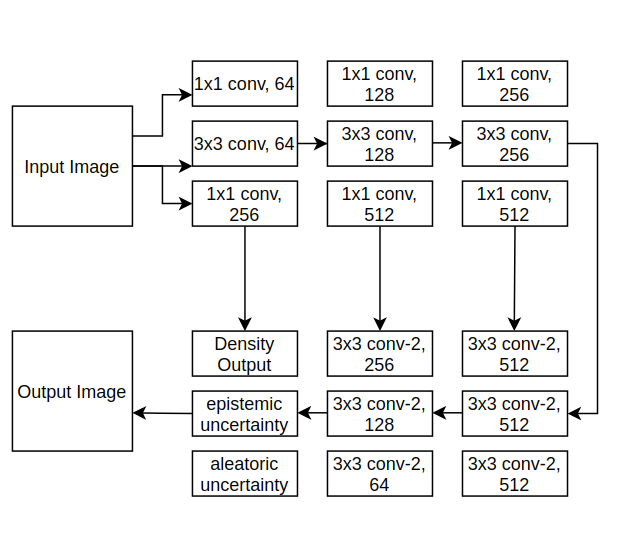}
\caption{Illustration of our proposed network architecture. In the diagram, 1$\times$1 and 3$\times$3 denote convolutional filter sizes, 64, 128, 256 indicate the receptive field sizes (feature channel depths), conv represents dilated convolutional layers used in the downsampling path, and conv-2 denotes transposed convolutional layers used for upsampling in the reconstruction path.}
\label{fig3}
\end{figure}

An overview of the architecture, including the layer-wise structure, is illustrated in Figure~\ref{fig3}.

This carefully designed architecture enables our model to effectively estimate crowd density while simultaneously quantifying uncertainty in a principled Bayesian framework.

\subsection{Optimization}

While training the network, the vanishing gradient problem showed up, ie weights of the connections were turning out to be zero. To alleviate this, instance normalization was used after both convolutional and transposed convolutional layers as defined below:

\begin{equation}
y={ReLU}\left(\sum_{i=0}^{d} w_{i} \cdot {ReLU}\left(\gamma_{i} \cdot \frac{x_{i}-\mu_{i}}{\sqrt{\sigma_{i}^{2}}+\epsilon}+\beta_{i}\right)+b\right)    
\end{equation}

Where $w$ and $b$ are the weight and bias terms of the convolution layer, $\gamma$ and $\beta$ are the weight and bias terms of the Instance Normalization layer, $\mu$ and $\sigma$ are the mean and variance of the input.

Previous works have used multi-column architecture \citep{zhang2016single} to deal with the various scales at which objects might be present in the image. The problem with these methods is that the number of columns gives a direct measure of the scale at which they can recognize individual objects. To tackle this, we propose a new technique to aggregate the filters with sizes 1$\times$1, 3$\times$3, and 5$\times$5. ReLU is applied after every convolutional and transposed convolutional layer. The filter branches make our network robust and can be extended by using more filters to tackle crowd counting in dense scenes. Our aggregation modules stacked on top of each other behave as ensembles, thus minimizing overfitting, which is a challenge with deep networks. The novel aggregation module used in our work is shown in \autoref{fig4}:

\begin{figure}[htp]
    \centering
    \includegraphics[width=8cm]{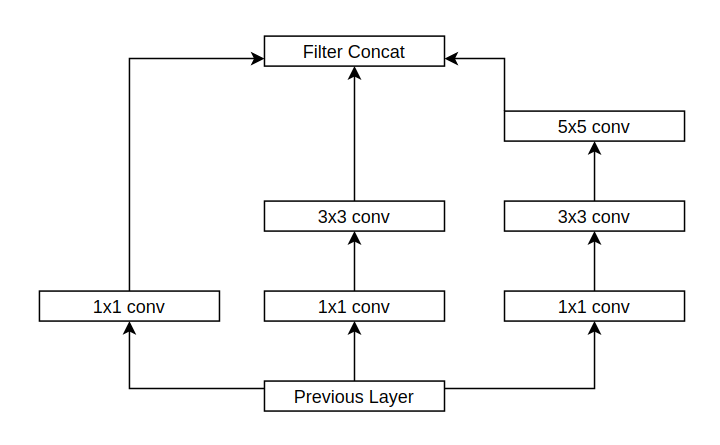}
    \caption{Illustration of our aggregation module.}
    \label{fig4}
\end{figure}

\subsection{Loss Function}

Most existing work uses pixel-wise Euclidean loss for training the network. This gives a measure of estimation error at the pixel level, which is defined below:

\begin{equation}
L_{E}=\frac{1}{N}\|F(X ,\theta)-Y\|_{}^{2}   
\end{equation}

where $\theta$ denotes a set of the network parameters, $N$ is the number of pixels in density maps, $X$ is the input image and $Y$ is the corresponding ground truth density map, $F(X, \theta)$ denotes the estimated density map. 

We also incorporate the SSIM index in our loss to measure the deviation of the prediction from the ground truth. The SSIM index is used in image quality assessment. It computes similarity between two images from three local statistics, i.e., mean, variance, and covariance. The range of SSIM values is from -1 to 1, and the SSIM is equal to 1 when the two images are identical. The SSIM index is defined in:

\begin{equation}
S S I M=\frac{\left(2 \mu_{F} \mu_{Y}+C_{1}\right)\left(2 \sigma_{F Y}+C_{2}\right)}{\left(\mu_{F}^{2}+\mu_{Y}^{2}+C_{1}\right)\left(\sigma_{F}^{2}+\sigma_{Y}^{2}+C_{2}\right)}
\end{equation}

where $C_{1}$ and $C_{2}$ are small constants to avoid division by zero. The next term of the loss function can be written by averaging over the integral, as shown below:

\begin{equation}
L_{S}=\frac{1}{N} \sum_{{x}} S S I M({x})    
\end{equation}

Where $N$ is the number of pixels in the density maps. $L_{S}$ gives a measure of the difference between the network predictions and ground truth. The final loss function, by adding the two terms, can be written as shown in \autoref{final_loss}:

\begin{equation}
L_{tot}=\alpha_{E} L_{E}+\alpha_{S} L_{S}   
\label{final_loss}
\end{equation}

where $\alpha_{E}$ and $\alpha_{S}$ are constants. In our experiments, we set both $\alpha_{E}$ and $\alpha_{S}$ as 0.5 to give equal weights to both the terms.

\subsection{Evaluation Metrics}

For crowd counting, the count error is measured by two metrics, Mean Absolute Error (MAE) and Mean Squared Error (MSE), which are commonly used for quantitative comparison. These metrics are defined as in \autoref{mae} and \autoref{mse}:

\begin{equation}
M A E=\frac{1}{N} \sum_{i=1}^{N}\left|C_{i}-C_{i}^{G T}\right|    
\label{mae}
\end{equation}

\begin{equation}
M S E=\sqrt{\frac{1}{N} \sum_{i}^{N}\left|C_{i}-C_{i}^{G T}\right|^{2}}  
\label{mse}
\end{equation}

Where $N$ is the number of test samples, $C_{i}$ and $C_{i}^{G T}$ are the estimated and ground truth count corresponding to the $i^{th}$ sample, which is given by the integration of the density map. MAE shows the accuracy of the predicted result, while MSE measures the robustness of the prediction.

\subsection{Uncertainty Estimation}

In predictive modeling, especially for safety-critical tasks like crowd counting in highly congested scenes, quantifying uncertainty is essential. Uncertainty provides a measure of confidence in model predictions and is typically categorized into two main types: epistemic uncertainty and aleatoric uncertainty.

Epistemic uncertainty (also known as model uncertainty) arises due to the lack of sufficient knowledge or data. It reflects uncertainty in the model parameters and can, in theory, be reduced by collecting more diverse and representative training data. Epistemic uncertainty is especially prominent in regions of the input space that the model has not seen during training.

Aleatoric uncertainty (also called data uncertainty) stems from inherent noise in the observations—for example, occlusion, scale ambiguity, low-resolution imagery, poor lighting, or clutter. This form of uncertainty cannot be eliminated by gathering more data, as it is intrinsic to the data-generating process.

To capture epistemic uncertainty, we leverage Bayesian Neural Networks (BNNs), where the weights of the network are treated as distributions rather than deterministic point estimates. This is achieved by placing a prior distribution over the weights and approximating the posterior using Monte Carlo dropout technique. This probabilistic treatment allows the model to express uncertainty in the learned representations, especially useful in out-of-distribution or ambiguous regions.

On the other hand, aleatoric uncertainty is modeled directly as a learnable component of the network’s output, allowing the model to predict heteroscedastic noise—i.e., noise that varies across input samples. To simultaneously learn the predictive mean and variance, we define a loss function that captures both types of uncertainty. The loss function used for training our network is depicted below:

\begin{equation}
\mathcal{L}(\theta)=\frac{1}{D} \sum_{i} \frac{1}{2 \sigma^{2}}\left\|y_{i}-\hat{y}_{i}\right\|^{2}+\frac{1}{2} \log \sigma^{2} 
\end{equation}

where $y_{i}$ is the $i^{th}$ pixel of the output density $y$ corresponding to input $x$ and $D$ is the number of output pixels. Note that the observation noise $\sigma^{2}$ captures how much noise is present in the outputs, and it stays constant for all data points. 

\subsection{Algorithm}

Let input images be denoted by $\left\{x_{n}\right\}_{n=1}^{N}$ and ground truth images by $\left\{y_{n}\right\}_{n=1}^{N}$. The trainable parameters for the network are denoted by $\theta, \phi$, which are obtained from a uniform distribution $\{1, \ldots, K\}$. The model parameters are denoted by $\theta$ for the shared backbone and $\phi$ for task-specific heads (density map, epistemic uncertainty, aleatoric uncertainty). To capture the predictive uncertainty, we model $\theta$ and $\phi$ as random variables and approximate their distributions by sampling from a uniform prior over $K$ different weight samples during training.

The complete algorithm used in our work is shown below:

\begin{algorithm}
\SetAlgoLined
 Require: Input images $\left\{x_{n}\right\}_{n=1}^{N},$ GT images $\left\{y_{n}\right\}_{n=1}^{N}$\\
 Initialize parameters $\theta, \phi$\\
 \For{each epoch}{
 \For{n = 1 to N}{
 Sample $\theta, \phi \sim$ Uniform $\{1, \ldots, K\}$
 
 Compute predictions $\left[{y}_{n}\right]=f_{{\theta}_{k}}\left(x_{n}\right)$
 
 Calculate loss: ${L}(\theta)=\frac{1}{D} \sum_{i} \frac{1}{2 \sigma^{2}}\left\|y_{i}-\hat{y}_{i}\right\|^{2}+\frac{1}{2} \log \sigma^{2}$
 
 Update $\theta_{k}$ using gradient descent $\frac{d {L}\left(\theta_{k}\right)}{d \theta_{k}}$
 
 }
 
 }
 \caption{Bayesian Multi-Scale Neural Network for Crowd Counting}
\end{algorithm}

The uniform sampling from $K$ different parameterizations introduces stochasticity to emulate Bayesian posterior sampling. The loss function jointly minimizes prediction error and learns to predict aleatoric uncertainty. Epistemic uncertainty is captured through weight sampling at inference time by averaging multiple forward passes.

\section{Experimental Results and Analysis}

\subsection{Quantitative Results}

To evaluate the effectiveness of our proposed Bayesian multi-scale network for crowd counting, we conducted extensive experiments on three benchmark datasets: ShanghaiTech, UCF-CC 50, and UCF-QNRF. Our model consistently achieves the lowest Mean Absolute Error (MAE) and Mean Squared Error (MSE) across all datasets, demonstrating both accuracy and robustness. In addition, we report the number of trainable parameters to show that our method is not only accurate but also highly efficient.

The ShanghaiTech dataset consists of two subsets: Part A with dense crowd scenes and Part B with sparse crowds. As shown in Table \ref{sample-table2}, our method outperforms most previous state-of-the-art approaches in terms of both MAE and MSE on both subsets. Compared to CSRNet, CP-CNN, and Switch-CNN, our model achieves better accuracy while maintaining a significantly smaller parameter footprint (see Table \ref{sample-table1}).

\begin{table*}
\caption{Comparison with state-of-the-art methods on ShanghaiTech dataset (lower is better). Left: Part A, Right: Part B}
\label{sample-table2}
\centering
\begin{tabular}{lllll}
\toprule
\textbf{Method} &\textbf{MAE (A)} &\textbf{MSE (A)} &\textbf{MAE (B)} &\textbf{MSE (B)}\\
\midrule
Zhang et al. \citep{zhang2015cross} &181.8 &277.7 &32.0 &49.8\\
MCNN \citep{zhang2016single} &110.2 &173.2 &26.4 &41.3\\
Cascaded-MTL \citep{sindagi2017cnn} &101.3 &152.4 &20.0 &31.1\\
Switch-CNN \citep{sam2017switching} &90.4 &135.0 &21.6 &33.4\\
CP-CNN \citep{sindagi2017generating} &73.6 &106.4 &20.1 &30.1\\
CSRNet \citep{li2018csrnet} &68.2 &115.0 &10.6 &16.0\\
SANet \citep{cao2018scale} &67.0 &104.5 &8.4 &13.6\\
SFCN \citep{wang2019learning} &64.8 &107.5 &7.6 &13.0\\
CAN \citep{liu2019context} &\textbf{62.3} &100.0 &7.8 &12.2\\
DUBNet \citep{oh2020crowd} &64.6 &106.8 &7.7 &12.5\\
\textbf{Ours} &63.2 &\textbf{95.6} &\textbf{7.3} &\textbf{10.6}\\
\bottomrule
\end{tabular}
\end{table*}

UCF-CC 50 is a highly challenging dataset due to its extremely limited size (only 50 images) and wide density variation. Table \ref{sample-table3} shows that our method achieves the lowest MSE while maintaining a competitive MAE, outperforming several high-capacity models such as SFCN and DUBNet. The balance between accuracy and generalization under extreme data scarcity demonstrates the strength of our Bayesian modeling approach.

\begin{table}[hbt!]
\caption{Comparison with state-of-the-art methods on UCF-CC 50 dataset (lower is better)}
\label{sample-table3}
\centering
\begin{tabular}{lll}
\toprule
\textbf{Method} &\textbf{MAE} &\textbf{MSE}\\
\midrule
MCNN \citep{zeng2017multi} &377.6 &509.1\\
Cascaded-MTL \citep{sindagi2017cnn} &322.8 &397.9\\
Switch-CNN \citep{sam2017switching} &318.1 &439.2\\
D-ConvNet \citep{shi2018crowd} &288.4 &404.7\\
L2R \citep{wan2019residual} &279.6 &388.9 \\
CSRNet \citep{li2018csrnet} &266.1 &397.5\\
ic-CNN \citep{ranjan2018iterative} &260.9 &365.5\\
SANet \citep{cao2018scale} &258.4 &334.9\\
SFCN \citep{wang2019learning} &214.2 &318.2\\
CAN \citep{liu2019context} &\textbf{212.2} &243.7\\
DUBNet \citep{oh2020crowd} &243.8 &329.3\\
\textbf{Ours} &216.7 &\textbf{225.1}\\
\bottomrule
\end{tabular}
\end{table}

UCF-QNRF is one of the largest and most diverse crowd counting datasets, with highly congested scenes and large image resolution. Table \ref{sample-table4} highlights that our method achieves the best MSE performance and remains competitive in MAE compared to other strong baselines like DUBNet and CAN. Our improved uncertainty modeling helps in better generalization across such diverse scenes.

\begin{table}[hbt!]
\caption{Comparison with state-of-the-art methods on UCF-QNRF dataset (lower is better)}
\label{sample-table4}
\centering
\begin{tabular}{lll}
\toprule
\textbf{Method} &\textbf{MAE} &\textbf{MSE}\\
\midrule
MCNN \citep{zeng2017multi} &277 &426\\
Cascaded-MTL \citep{sindagi2017cnn} &252 &514\\
Switch-CNN \citep{sam2017switching} &228 &445\\
CSRNet \citep{li2018csrnet} &135.5 &207.4\\
SFCN \citep{wang2019learning} &\textbf{102.0} &171.4\\
CAN \citep{liu2019context} &107 &183\\
DUBNet \citep{oh2020crowd} &105.6 &180.5\\
\textbf{Ours} &106.7 &\textbf{165.1}\\
\bottomrule
\end{tabular}
\end{table}

Beyond accuracy, our model is designed to be lightweight and computationally efficient. As shown in Table \ref{sample-table1}, our method achieves state-of-the-art performance using only 0.24 million parameters, which is significantly fewer than even the most compact prior works like SANet (0.91M). This makes our architecture well-suited for deployment on edge devices and real-time applications.

\begin{table*}
\caption{Comparison of model size in terms of parameter count (in millions). Lower is better.}
\label{sample-table1}
\centering
\begin{tabular}{llllll}
\toprule
\textbf{Method} &Switch-CNN \citep{sam2017switching} &CP-CNN \citep{sindagi2017generating} &CSRNet \citep{li2018csrnet} &SANet \citep{cao2018scale} &\textbf{Ours}\\
\midrule
\textbf{Parameters (M)} &15.11 &68.4 &16.26 &0.91 &\textbf{0.24}\\
\bottomrule
\end{tabular}
\end{table*}

\subsection{Qualitative Results and Uncertainty Analysis}

\begin{figure}
\centering
\includegraphics[width=8cm]{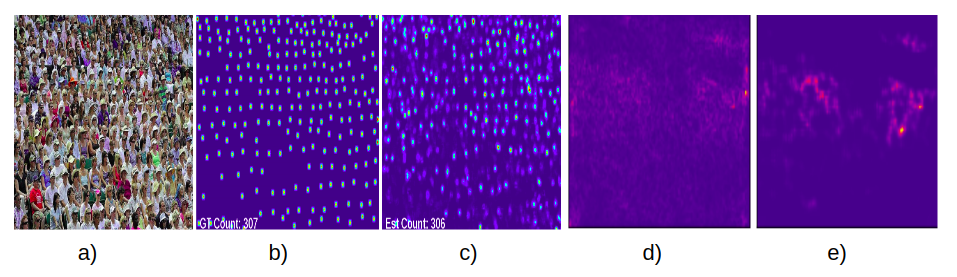}
\caption{Qualitative results on the ShanghaiTech dataset. Each row shows: (a) Input image, (b) Ground truth density map, (c) Predicted density map, (d) Epistemic uncertainty, and (e) Aleatoric uncertainty.}
\label{fig1}
\end{figure}

\begin{figure}
\centering
\includegraphics[width=8cm]{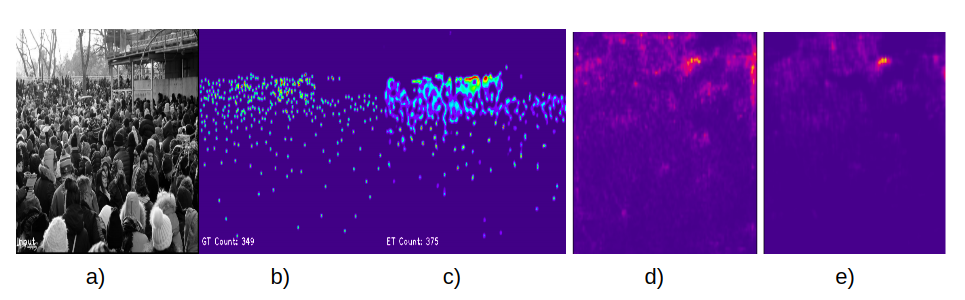}
\caption{Qualitative results on the UCF-QNRF dataset. Each row shows: (a) Input image, (b) Ground truth density map, (c) Predicted density map, (d) Epistemic uncertainty, and (e) Aleatoric uncertainty.}
\label{fig2}
\end{figure}

Figures \ref{fig1} and \ref{fig2} present qualitative results of our proposed method on representative samples from the ShanghaiTech and UCF-QNRF datasets, respectively. Each row displays a crowd image from the test set alongside five visualizations: the input image, the ground-truth density map, the predicted density map, and the corresponding epistemic and aleatoric uncertainty maps.

The epistemic uncertainty reflects the model's uncertainty due to limited training data or model capacity. It is learned through multiple stochastic forward passes and captures the spread of the model's predictions. On the other hand, the aleatoric uncertainty represents the inherent noise and ambiguity in the input data—such as motion blur, low resolution, or occlusion—which cannot be reduced even with more data.

In both datasets, we observe the following:

\begin{itemize}
    
\item Higher uncertainty in regions of dense crowds: In areas with high object overlap or extreme perspective distortion, both epistemic and aleatoric uncertainty values are notably elevated. This is expected, as accurately estimating density in such regions is inherently more difficult.

\item Correlation between uncertainties: There is a visible spatial alignment between regions of high epistemic and aleatoric uncertainty, indicating that ambiguous regions in the image (e.g., occluded or poorly lit people) challenge the model both from a data and modeling perspective.

\item Sharper epistemic patterns in sparse areas: In low-density regions, the epistemic uncertainty tends to capture specific regions where the model is unsure about the existence of crowd presence, while aleatoric uncertainty remains low—highlighting model doubt rather than data ambiguity.

\item The color intensity in the uncertainty maps—particularly the red regions—correlates with the degree of uncertainty: more red denotes higher uncertainty. This visual feedback can be crucial in real-world applications where knowing the model's confidence is as important as the prediction itself.

In summary, the proposed method not only provides accurate density maps but also delivers meaningful uncertainty quantification that helps interpret the reliability of its outputs, especially under challenging crowd scenes.

\end{itemize}

\section{Conclusions}

In this work, we proposed a novel deep learning framework for crowd counting that integrates accurate density estimation with robust uncertainty quantification. The architecture is built upon a ResNet-based feature extractor, augmented with dilated convolutions in the downsampling path to preserve spatial resolution and capture multi-scale context. The upsampling path leverages transposed convolutions, while skip connections between corresponding encoder and decoder layers promote effective feature reuse, mitigate vanishing gradients, and help prevent overfitting. To enhance prediction robustness, we introduced a feature aggregation module that facilitates rich semantic fusion across different levels of the network. Furthermore, the network branches into three output heads: a density map for crowd count estimation, and two auxiliary heads to estimate epistemic and aleatoric uncertainty, thereby making the model’s predictions more interpretable and trustworthy. We also detailed the Bayesian learning framework employed to model epistemic uncertainty via variational weight sampling and used a log-likelihood-based loss function to capture aleatoric noise. A complete training algorithm was provided to demonstrate how uncertainty-aware optimization is implemented end-to-end. Experimental evaluations on three benchmark datasets—ShanghaiTech, UCF-CC 50, and UCF-QNRF—demonstrate that our model achieves state-of-the-art performance, consistently outperforming prior methods in both MSE and MAE metrics. Additionally, our model achieves this with a significantly lower parameter count, showcasing its efficiency and scalability. Importantly, the integration of uncertainty modeling addresses the black-box nature of traditional deep neural networks by providing pixel-wise estimates of prediction confidence. This capability is especially crucial for deployment in high-stakes real-world applications such as public safety, event monitoring, and urban planning.

\nocite{*}
{
    \small
    \bibliographystyle{ieeenat_fullname}
    \bibliography{main}
}

\end{document}